%% file: root.tex
\documentclass[letterpaper, 10 pt, conference]{ieeeconf}
\IEEEoverridecommandlockouts                 
\usepackage{amsmath,amssymb,amsfonts}
\usepackage{algorithmic}
\usepackage{graphicx}
\usepackage{textcomp}
\usepackage{xcolor}
\usepackage{caption}
\usepackage{subcaption}
\usepackage{svg}
\usepackage{balance}

\title{\LARGE \bf Good Enough to Learn: LLM-based Anomaly Detection in ECU Logs without Reliable Labels
}

\author{Bogdan Bogdan$^{1,*}$, Arina Cazacu$^{1,*}$, Laura Vasilie$^{1,*}$
\thanks{*All authors contributed equally.}
\thanks{$^{1}$Dept. of AI, Big Data and Cloud
Porsche Engineering Romania SRL Cluj-Napoca, Romania, Emails: {\tt\small first-name.last-name@porsche-engineering.ro}}%
}

\begin{document}

\maketitle
\thispagestyle{empty}
\pagestyle{empty}

\begin{abstract}

Anomaly detection often relies on supervised or clustering approaches, with limited success in specialized domains like automotive communication systems where scalable solutions are essential. We propose a novel decoder-only Large Language Model (LLM) to detect anomalies in Electronic Control Unit (ECU) communication logs. Our approach addresses two key challenges: the lack of LLMs tailored for ECU communication and the complexity of inconsistent ground truth data. By learning from UDP communication logs, we formulate anomaly detection simply as identifying deviations in time from normal behavior. We introduce an entropy regularization technique that increases model's uncertainty in known anomalies while maintaining consistency in similar scenarios. Our solution offers three novelties: a decoder-only anomaly detection architecture, a way to handle inconsistent labeling, and an adaptable LLM for different ECU communication use cases. By leveraging the generative capabilities of decoder-only models, we present a new technique that addresses the high cost and error-prone nature of manual labeling through a more scalable system that is able to learn from a minimal set of examples, while improving detection accuracy in complex communication environments.

\end{abstract}

\input{chapters/introduction}

\input{chapters/related_work}

\input{chapters/methodology}
\input{chapters/results}

\input{chapters/conclusion}

\bibliographystyle{IEEEtran}
\bibliography{references}

\end{document}

%% file: chapters/introduction.tex
\section{Introduction}
 In a vehicle, a substantial volume of data is transmitted between sensors and small computing units, in order to ensure a smooth functioning of various systems in the car. These units, often called ECUs (Electronic Control Units), communicate with each other via multiple protocol types (e.g UDP, TCP, Ethernet, CAN). We call such a collection of messages a communication log or trace. Every time an update is made to an ECU, the entire system has to be intensively tested to see whether the communication still takes place as expected or issues such as delays, incomplete messages or even total absences occur.
 
Determining whether a given message transmitted over a bus is an anomaly can be quite challenging, because several factors must be considered, including the time since its last occurrence, the payload, the system state, the queuing and routing strategy and the signal frequency, among others. Also, as systems evolve and accumulate new functionalities, they inherently become increasingly complex. A classical trace analysis approach consists in implementing all these rules using highly coupled conditional statements, which in such a complex system might be error prone and very difficult to validate or maintain. 
 
 Therefore, our proposal is to train a Large Language Model (LLM) that learns the ECUs communication language and then fine-tune it for detecting anomalies. This way, the testing strategy becomes more robust and scalable, LLMs being a great choice when it comes to scalability and generalization. Moreover, it seems like there are no other decoder-only models applied for the anomaly detection use case or in general LLMs trained on predicting ECU communication logs in the current literature, so novelty also plays a strong part in our motivation.

 The main challenge that we had to overcome was that the classical anomaly detection implementation based on rules did not yield reliable labels, so directly using them as ground truth (GT) in the fine-tuning step was not a good option. This led us to taking the Open Word Assumption (OWA) principle into consideration for some of the apparently non-anomalous logs: just because something is not known to be true does not automatically mean it is false. Thus, the fine-tuning step is developed based on OWA and we use the inconsistent annotations in our advantage. Also, this only requires a small and inexpensive labeled dataset.

 The goal of this research is to present a method for training a Large Language Model to detect anomalies based on inconsistently labeled data, with the possibility of generalization in future work as a framework for working with unreliable annotations in other use cases as well. The method includes two main steps: pre-training the LLM on general ECU communication logs for next token prediction task and then fine-tuning on anomaly detection with inconsistent labels. For this proof of concept, we extracted data from the vehicle with only one ECU as source while the destination may vary, and the communication takes place via Ethernet, based on UDP, the focus being on detecting time related anomalies (delayed or missing messages, further referred to as cycle time anomalies). This analysis is intended for offline execution on data gathered from the vehicle. Our contribution is three-fold: decoder-only anomaly detection, training on inconsistent labels and an LLM pre-trained on ECU communication data that can be further adapted for other use cases. 
All these equally enable us to build an original approach for finding anomalies without having reliable ground truth: we guide the model to focus more on the few known anomalies, while we also prove that it is then capable of generalizing and correctly identifying others as well.

%% file: chapters/related_work.tex
\section{Related Work}

As summarized in a systematic literature review \cite{su2024largelanguagemodelsforecasting} published recently (February 2024), the most popular and well performing choice when it comes to LLMs on anomaly detection is BERT \cite{devlin2019bertpretrainingdeepbidirectional}, due to its bidirectional approach and only a small labeled dataset being needed to achieve better results than state-of-the-art work. While BERT is built on Transformer encoder architecture,  this paper proposes a decoder-only model for detecting anomalies due to its superior handling of sequential dependencies making it inherently more suitable for detecting cyclic patterns and more scalable, allowing it to capture distant events over longer context windows. A couple of well known specific models are LogBERT \cite{guo2021logbertloganomalydetection}, followed by LanoBERT \cite{lee2023lanobertloganomalydetection} which does not include abnormal data in training, hence requiring a smaller amount of such annotations only when testing. Since labeling inconsistencies might be present in all of our data, this strategy was not applicable. 

If we refer to LLMs for ECU communication logs, CAN-BERT \cite{alkhatib2022canbertitcontrollerarea} is a model created in a very similar setting with the one in this paper, focused on detecting intrusions on CAN (Controller Area Network) data, instead of Ethernet/UDP. These attacks are also reflected as deviations in time and they obtain good results on benchmark datasets with labeled anomalies.

Regarding inconsistent labeling, there is research based on weakly supervised anomaly detection methods that suggest aligning the input data with its corresponding occurrence rules via two encoders and computing a joint loss between the two \cite{zhao2024weaklysupervisedanomalydetection}. In our setup, the rules are not represented in natural language, therefore collection and translation would imply an extra step.  Another notable approach is called FATE \cite{das2023fewshotanomalydetectiontext} and it relies on deviating the known anomalous inputs as much as possible from the initial reference distribution through scores. While the main idea would fit for our context as well, FATE is a few-shot framework for text data. 

All things considered, from the three main contributions that we presented at the end of the previous chapter, none is specifically tackled in literature so far, to the best of our knowledge. 

%% file: chapters/methodology.tex
\section{Methodology}

Our proposal follows a two-stage approach:

\begin{enumerate}
    \item In the first stage, we pre-train a model such that it learns the general ECU communication protocol.
    \item In the second stage, we fine-tune the pre-trained model for the anomaly detection use case.
\end{enumerate}

\subsection{Data Overview}

The raw data consists of multiple traces of messages collected during vehicle testing between the main high-performance computing platforms (HCP) and multiple ECUs. Data acquisition is performed using a vehicle spy logger placed on the bus to analyze message traffic from multiple networks. The chosen ECU communication protocol is UDP-based. Each trace contains multiple messages called protocol data units (PDUs). These are packets handled by the network layer, according to the OSI model.

\begin{figure}[htbp]
    \centering
    \includegraphics[width=1.0\linewidth]{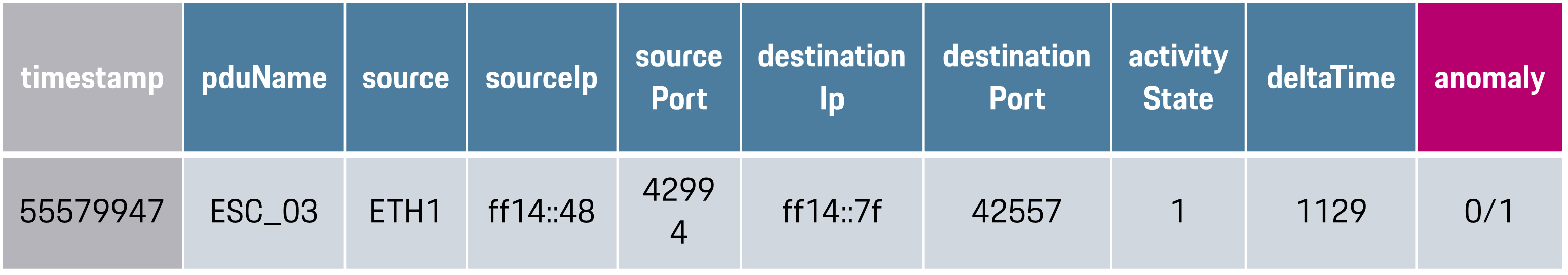}
    \caption{General structure of a message}
    \label{fig:pdu}
\end{figure}

The general structure of a message is presented in Figure \ref{fig:pdu}. During data acquisition, collected messages contain multiple types of fields. However, we select only those relevant to our task, which are:

\begin{enumerate}
    \item \textit{Name}: Identifies the signal type sent by the message.
    \item \textit{Source}: The main HCP and protocol responsible with the communication with multiple ECUs. 
    \item \textit{Source IP Address and Port}: Logical address of the sender and its associated port.
    \item \textit{Destination IP Address and Port}: Logical address of the receiver and its associated port.
    \item \textit{Timestamp}: Absolute value measured in microseconds denoting the time when the message was sent.
    \item \textit{Delta Time}: Relative value measured in microseconds denoting the amount of time since the previous occurrence of a message for the same PDU.
    \item \textit{Activity State}: Indicates the current operation mode of the vehicle: normal operation, startup-shutdown, not initialized.
\end{enumerate}

\begin{figure}[htbp]
    \centering
    \includegraphics[width=1.0\linewidth]{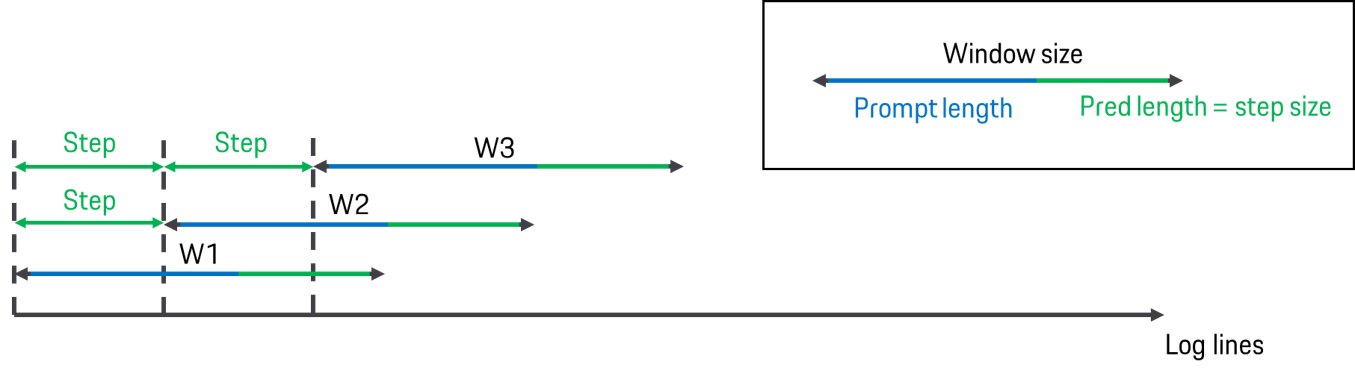}
    \caption{Sliding window approach}
    \label{fig:sliding-window}
\end{figure}

The methodology leverages decoder-only Transformer-based models. Because their context is limited in size, we create windows of messages following a sliding window manner over the traces. This approach is shown in Figure \ref{fig:sliding-window}. For each log file, we start at the beginning and select the first $W$ messages, where $W$ represents the window size. This window is split in two parts: the prompt and the part to be predicted by the model of length $P$. The next window is selected by moving $P$ steps further in the log file. The sliding window approach makes sure the data is fully used, as one sequence of messages can be considered for both context and prediction.

\subsection{Tokenization Strategy}

The ECU communication protocol can be seen as the "language" spoken by HCPs and ECUs. However, compared to natural language, it has a different vocabulary and grammar, and follows strict rules (e.g., synchronous PDUs have specific cycle times). So we have to adapt the tokenization strategy based on the ECU communication "language".

Our main tokenization strategy is based on Byte Pair Encoding (BPE), which was initially developed for text compression \cite{sennrich2016neuralmachinetranslationrare}. Its first usage in natural language processing was done by OpenAI, as the tokenization strategy for pre-training the GPT-2 model \cite{Radford2019LanguageMA}.

This algorithm replaces the most frequent pair of characters in a sequence with a new unique symbol, in an iterative way. Compared to strings, numerical features, such as $deltaTime$, are tokenized digit by digit. It proved to effectively balance the trade-off between vocabulary size and representation power. 

\begin{figure}[htbp]
    \centering
    \includegraphics[width=1.0\linewidth]{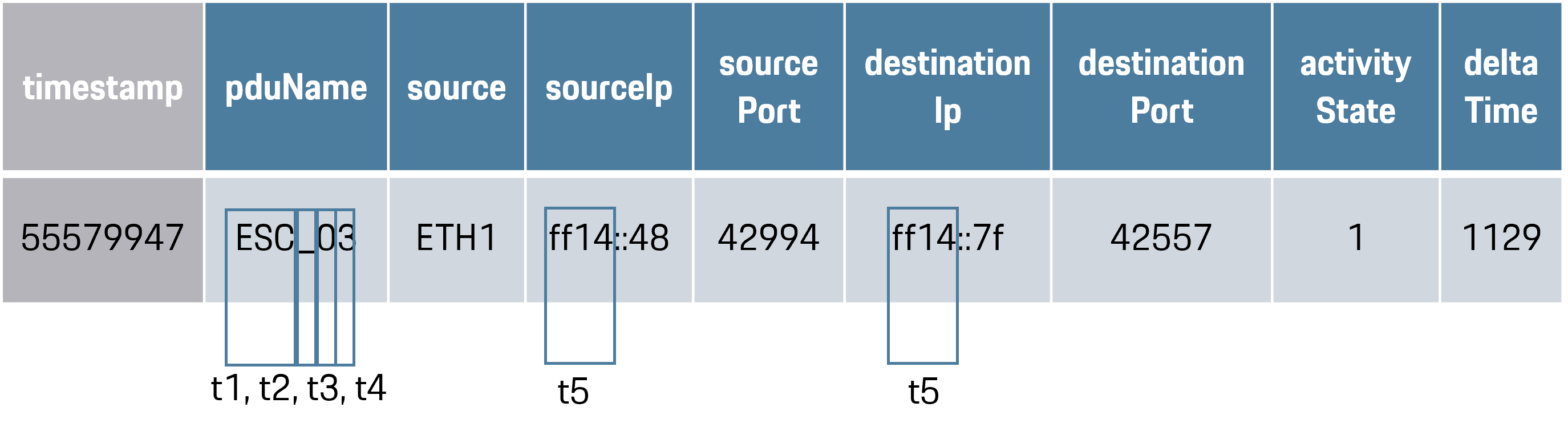}
    \caption{Custom Byte Pair Encoding}
    \label{fig:bpe}
\end{figure}

Figure \ref{fig:bpe} shows how byte pair encoding is used on the given data format. In the end, we obtain a BPE-based tokenizer with a vocabulary of size 408.

\subsection{Pre-Training}
In the pre-training stage, we initialize a new model instance based on $Qwen2$ architecture, developed by Qwen team from Alibaba Cloud \cite{yang2024qwen2technicalreport}. It is a family of open-source, decoder-only Transformer models with competitive performance compared to proprietary models on multiple benchmarks, ranging from language understanding to mathematics \cite{yang2024qwen2technicalreport}. 

In the field of large language models, the general approach is to pre-train a model on large text corpus and then fine-tune it for a specific task. Because pre-training can be quite expensive, in practice, there are open-source pre-trained models that could be directly fine-tuned on requested use cases. 

However, for the given problem, the "language" spoken by ECUs is different from natural language. As the data domain is different from natural language processing, we directly train a model from scratch on ECU communication ”language”. \cite{DBLP:journals/corr/abs-2004-10964}

The pre-training phase requires the model to be trained for the next token prediction task, using cross-entropy loss function, over a vocabulary of size $V$:

\begin{equation}
    L_{pretraining} = -\sum_{i=1}^Vy_{i}\log(\hat{y}_{i})
\end{equation}

\subsection{Fine-Tuning}

In the fine-tuning phase we specialize the pre-trained model on the anomaly detection task using LoRA \cite{hu2021loralowrankadaptationlarge}. Normally one can build a classifier in a supervised or semi-supervised setting to classify each line to be anomalous or not, however, as we are working with inconsistent labels we do not know the meaning of the negative class so the question arises in the open world assumption: What can we do if we encounter an unlabeled anomaly in the dataset?

The solution is to train with NTP on data containing also anomalies and then make the model more uncertain in tokens with known anomalies. The intuition is that similar tokens will behave in the same way. This makes our solution more attractive as it does not require any special pre-processing on the pre-training data.

\begin{figure}[htbp]
\centering
\includegraphics[width=0.45\linewidth]{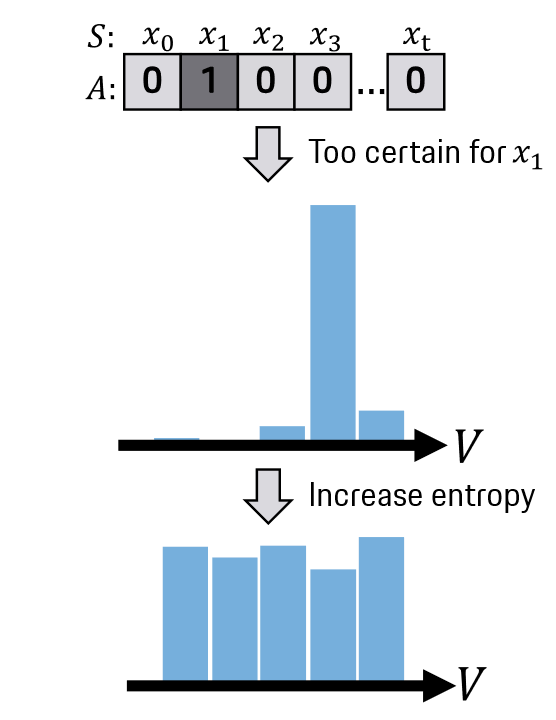}
\caption{Effect of entropy regularization on probability distribution of tokens with anomalies.}
\label{fig:entropy_reg}
\end{figure}

To increase uncertainty in tokens with anomalies we propose an entropy regularizer that maximizes entropy in anomaly tokens. This process is illustrated in Figure \ref{fig:entropy_reg}.

Formally, given a sequence of tokens $\mathcal{S} = (x_0, x_1, ..., x_t)$ with its corresponding anomaly binary mask given by $\mathcal{A} = (\hat{a}_0, \hat{a}_1, ..., \hat{a}_t)$ the entropy for a decoder-only LLM parameterized by $\theta$ is defined as:

\begin{equation}
    H(\mathcal{S}, \mathcal{A}|\theta) = - \sum_{i=1}^{V} \log p_{\theta}(x_i | x_{<i}) \cdot \hat{a}_i
\label{eq:entropy_decoder}
\end{equation}

Equation $PPL(\mathcal{S}, \cdot |\theta) = e^{H(\mathcal{S}, \cdot |\theta)}$ tells us that maximizing the entropy has also the effect of maximizing the perplexity as perplexity is the exponential of entropy, a frequently used metric for evaluating large language models and one of the metrics used in this work to detect anomalies.

Entropy regularization is suitable for fine-tuning any LLM architecture, either encoder-only, decoder-only or encoder-decoder transformers making it a general technique to deal with label inconsistency at scale. Another advantage is that due to its simplicity it can be coupled with other objective functions for a given use case as it only impacts the final probability distribution over the vocabulary. 

Note that the final objective for the fine-tuning phase is the combination between the cross-entropy loss for next token prediction and the entropy regularizer weighted by a factor $\alpha$ to control its influence:

\begin{equation}
    L = - \sum_{i=1}^{V} y_i \log(\hat{y}_i) - \alpha H(\mathcal{S}, \mathcal{A}|\theta)
\label{eq:final_loss}
\end{equation}

The loss in Equation \ref{eq:final_loss} is expressed for one sample $\mathcal{S}$ with shifted targets $\hat{y}$ and operates over the vocabulary $V$. The cross-entropy component makes sure no knowledge acquired during the pre-trained phase is lost while the regularizer addresses the label inconsistency issue.

Now that we have a model that understands both UDP communication and where possible inconsistencies might appear in the sequence, the raw outputs of the model can be directly used to perform anomaly detection. It can be seen that we can now decouple the raw predictions of the model from the formulation of some anomaly detection metrics. Finally, these metrics can simply be compared with predefined thresholds to arrive at $0/1$ anomaly labels. In what follows we propose two metrics for detecting anomalies with a decoder-only LLM.

\subsection{Anomaly Detection Metrics}

To interpret the final model's predictions for the anomaly detection use case, we use two metrics: top-$k$ and perplexity.

\subsubsection{Top-$k$ Metric} It takes into account the top-$k$ candidates yielded by the model for the next token prediction. If the ground truth token is not found among the top-$k$ options, then this token is considered as being anomalous.

\subsubsection{Perplexity Metric} This metric is frequently used for evaluating large language models. It evaluates how confused the model is when predicting the next token, for the given context. If the perplexity value is greater than a fixed threshold, then the ground truth token is considered as being anomalous.

\subsection{Final Design}

\begin{figure}[htbp]
    \centering
    \includegraphics[width=1.0\linewidth]{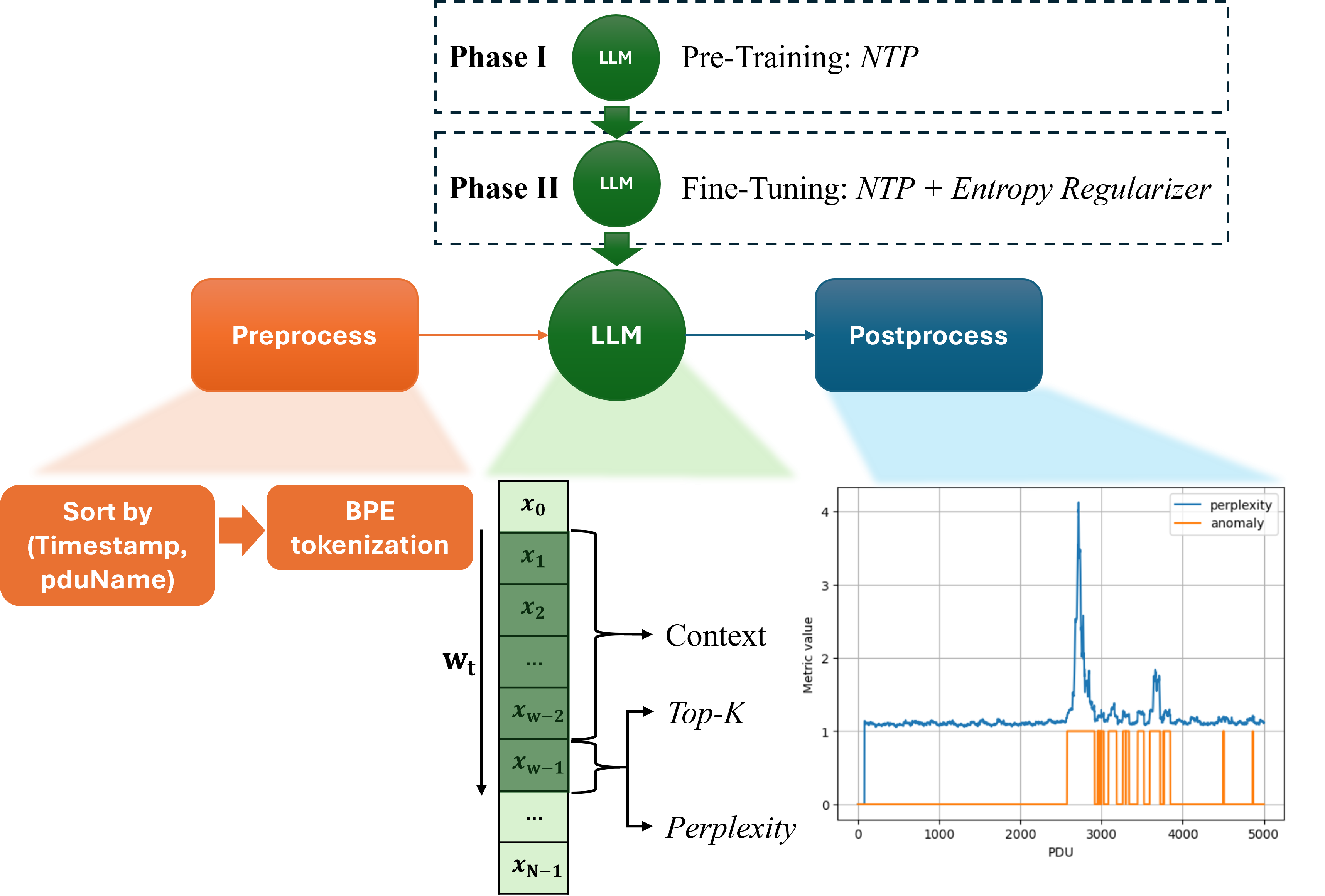}
    \caption{Final version of proposed solution}
    \label{fig:overview}
\end{figure}

Our proposed methodology is presented in Figure \ref{fig:overview}. A new model instance is initialized with random weights and pre-trained using NTP. Then it is further fine-tuned for the anomaly detection use case, using NTP and entropy regularization as objective function.

The final model version is used in the anomaly detection pipeline, in which:

\begin{enumerate}
    \item First step is to pre-process the data: this involves sorting by $timestamp$, $pduName$ and then tokenizing it.
    \item Second step is to run detection over windows $w_t$ from the traces of messages, resulting from the sliding window approach. In this way, we ensure a proper context of size $W$ - 1 for computing perplexity and top-$k$ metrics values for the last position of the window.
    \item In the final step, we apply filters for removing outliers and converting raw metric values into 0/1 classes. The filters only consider consecutive metric values above or below a fixed threshold, and remove the other values which violate this condition.
\end{enumerate}

%% file: chapters/results.tex
\section{Experiments and Results}

In this section we evaluate our approach quantitatively along different dimensions. We also perform experiments to select the best hyperparameters, models and objective functions for the task of anomaly detection. We conclude with visual results of anomalies detected by our solution.

\subsection{Multiple Models}\label{AA}
We use the architecture of $Qwen2$ \cite{yang2024qwen2technicalreport} randomly initialized for our vocabulary size of 408 tokens, resulting in a total of $400M$ parameters. We also experimented with the smallest model version of $Mamba$ ($130M$ parameters)\cite{gu2024mambalineartimesequencemodeling}, as it can deal with longer contexts, a desired property when working with ECU communication data. Both $Qwen2-0.4B$ and $Mamba-130M$ were trained on data with no overlap. For training these models we follow Chinchilla scaling law from Hoffman et. al \cite{hoffmann2022trainingcomputeoptimallargelanguage} and generate a total of $8B$ tokens. By employing $50\%$ overlap using our sliding window strategy from Figure \ref{fig:sliding-window}, we are able to accommodate the required number of tokens to train also a larger version of $Qwen$ with $1.3B$ parameters on a dataset of $26B$ tokens. All models are compared in terms of perplexity and accuracy in predicting the next token as shown in Table \ref{tab:results_models}, from which we can conclude that $Qwen$ has achieved the best accuracy. 

\begin{table}[htbp]
\caption{Comparison on different models on the pre-training phase}
\begin{center}
\begin{tabular}{|c|c|c|}
\hline
\textbf{Model} & \textbf{\textit{NTP Accuracy}}& \textbf{\textit{NTP Perplexity}} \\
\hline
Mamba-130M & 95.63 & 1.42  \\
\hline
Qwen-0.4B & 97.03 & 1.53  \\
\hline
Qwen-1.3B & 98.62 & 1.03  \\
\hline
\end{tabular}
\label{tab:results_models}
\end{center}
\end{table}

All models were pre-trained for 1 epoch on 4xNvidia A100 GPUs on UDP data including known and unidentified anomalies. The dataset is split into train, validation, and test sets, using a split ratio of $0.8/0.1/0.1$. We used a batch size of 24 with 8 gradient accumulation steps, gradient checkpointing, a learning rate of $5\mathrm{e}{-4}$ with linear decay over first $1000$ train steps.

\subsection{Multiple Tokenizers}

The main tokenizer from our solution is a Byte-Pair encoder (BPE) trained from scratch. We also experimented with different tokenizers of different types of granularity. Tokenization employs a trade-off between cycle time variation and scalability. Cycle time variations refer to different distances in time between two consecutive occurrences of the same PDU. Table \ref{tab:results_tokenizers} shows three experiments with three different tokenizers producing a different number of tokens per row. 

The Byte-Pair encoder produces an average of $50$ tokens per line including the separator token between lines. A window of 4096 tokens (approximately 82 lines) in case of BPE tokenizer captures $25\%$ of cycle time variations while the Row tokenizer is able to capture $75\%$ of these variations as more lines can fit into the same window. However a lower granularity of tokens per row does not scale with large datasets where there might be many unique combinations or it may result in many new unknown tokens when testing on novel data. Thus we are using BPE being the tokenizer with highest NTP performance on the pre-training phase.

\begin{table}[htbp]
\caption{Comparison on different tokenizers on the pre-training phase. The unique combination of all features produces 1 token per row. Grouping features into $pduName$ and the rest produces 2 tokens per row.}
\begin{center}
\begin{tabular}{|c|c|c|}
\hline
\textbf{Tokenizer} & \textbf{\textit{NTP Accuracy}}& \textbf{\textit{NTP Perplexity}} \\
\hline
Row (1 token/row) & 94.74 & 1.32  \\
\hline
Group (2 tokens/row) & 97.03 & 1.53  \\
\hline
BPE (multiple tokens/row) & 98.31 & 1.04  \\
\hline
\end{tabular}
\label{tab:results_tokenizers}
\end{center}
\end{table}

\subsection{Multiple Prompt Lengths}

We also analyzed the influence of the context length in predicting the next token. We found out that by increasing both the token granularity per line and the context length the NTP performance also improves as can be seen in Table \ref{tab:results_pl}. Theoretically, the higher bound for this value is $32K$ tokens (the context length of $Qwen2-0.5B$), while the lower bound should be one that captures all of the cycle time variations. In practice and in our case the training infrastructure limits us to a maximum prompt length of $4096$ thus our window can capture only $25\%$ of cycle time variations.
This limitation is mitigated by the fact that we are including explicitly in our features for each PDU the offset in time since the last occurrence ($deltaTime$ feature).

\begin{table}[htbp]
\caption{Comparison on different prompt lengths on the pre-training phase}
\begin{center}
\begin{tabular}{|c|c|c|}
\hline
\textbf{Prompt length} & \textbf{\textit{NTP Accuracy}}& \textbf{\textit{NTP Perplexity}} \\
\hline
256 & 92.13 & 1.55  \\
\hline
512 & 95.76 & 1.56  \\
\hline
1024 & 97.03 & 1.53  \\
\hline
4096 & 98.31 & 1.04  \\
\hline
\end{tabular}
\label{tab:results_pl}
\end{center}
\end{table}

\subsection{Multiple Objective Functions}

Apart from our main solution based on entropy we have also experimented with other fine-tuning approaches for the task of anomaly detection. 
So another experiment was to use a contrastive regularizer to account for OWA. More specifically, we are contrasting negative samples that are windows with at least one known anomaly from positive samples that are the same samples with those anomalies removed (likely non-anomalous). The intuition is that after using the contrastive regularizer they are not embedded close together anymore minimizing the impact of inconsistencies.

For the final solution we used the entropy regularizer which resulted in the highest recall, that measures how many of the labeled anomalies did we correctly identify. Apart from the recall, we need to also make sure the precision is high, as this is a good indicator of how many false positives the pipeline identifies. We prioritize recall as false positives might be actually correct due to the underlying inconsistent labeling process. If we go one step further and analyze the performance of the model in identifying important regions of anomalies we obtain a recall of $81\%$. Regions represent groups of consecutive anomalies where we consider that the prediction is correct if it lies in that region. Note that the final results captured in Table \ref{tab:results_ad} might be higher in reality as they reflect training on inconsistent anomaly labels.

\begin{table}[htbp]
\caption{Comparison on different objectives on the fine-tuning phase with $Qwen2-0.4B$ variant. We measure the precision and recall in detecting anomalies with respect to GT anomalies as marked by the traditional approach.}
\begin{center}
\begin{tabular}{|c|c|c|}
\hline
\textbf{Objective} & \textbf{\textit{Anomaly Recall}}& \textbf{\textit{Anomaly Precision}} \\
\hline
Contrastive & 0.35 & 0.63  \\
\hline
Entropy (lines) & 0.6 & 0.9  \\
\hline
Entropy (regions) & 0.81 & 0.81  \\
\hline
\end{tabular}
\label{tab:results_ad}
\end{center}
\end{table}

\subsection{Visual Results}
We also evaluate our detection pipeline qualitatively in two ways: by manually perturbing the communication and by comparing with GT cycle time anomalies marked by a traditional communication analysis system based on rules. 

\begin{figure}[htbp]
\centering
\begin{subfigure}[t]{0.49\linewidth}
    \centering
    \includegraphics[width=\linewidth]{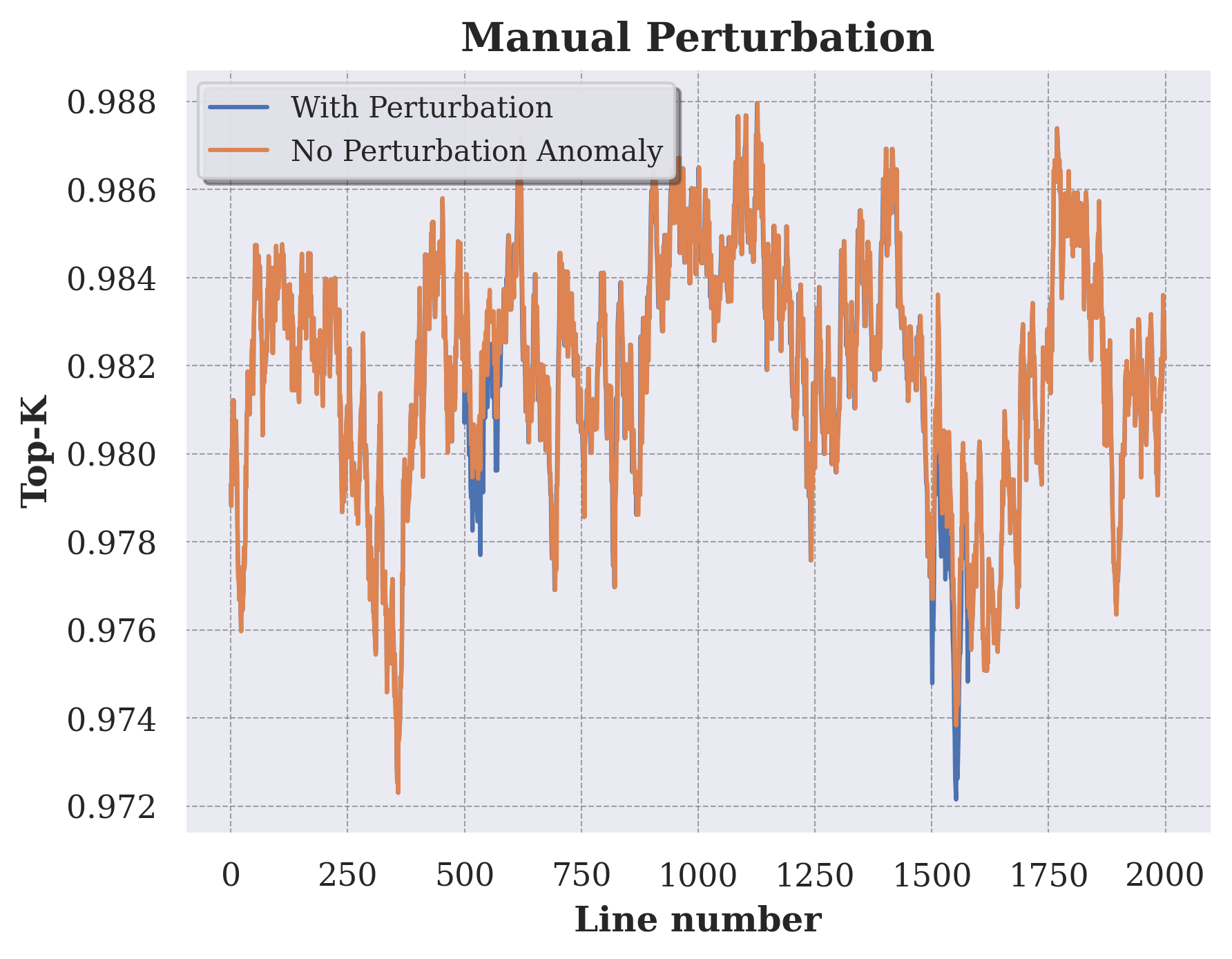}
    \caption{Detection using top-$k$ metric.}
    \label{fig:sensitivity_topk}
\end{subfigure}
\hfill
\begin{subfigure}[t]{0.49\linewidth}
    \centering
    \includegraphics[width=\linewidth]{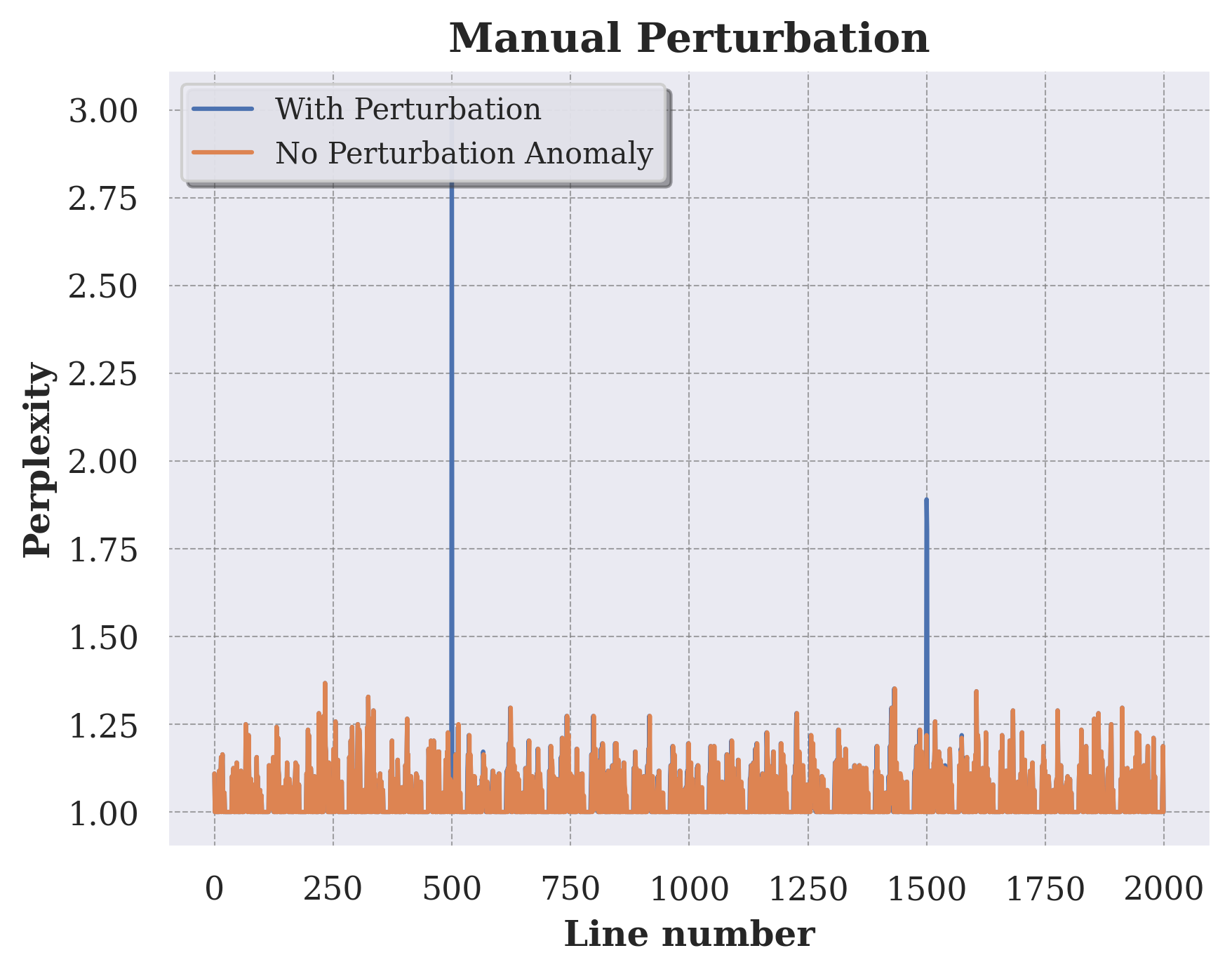}
    \caption{Detection using perplexity metric.}
    \label{fig:sensitivity_perplexity}
\end{subfigure}
\caption{Sensitivity analysis. Manually perturbing the normal operation sequence increases the model uncertainty: top-$k$ decreases (a) while perplexity (b) increases.}
\label{fig:sensitivity}
\end{figure}

Manual perturbation was performed on the base model, the artifact of phase 1 from our two-phase training methodology, and it involves manually moving a log line a few lines into the future simulating a time delay in microseconds. This measures the raw capability of the model to detect a delay in the normal communication sequence. In Figure \ref{fig:sensitivity} we show an example of a log file with 2000 lines where the model is sensitive for out-of-sequence events. Both perplexity and top-$k$ metrics begin to degrade in time instances (highlighted in blue) when events temporarily disappear (e.g., line 500) from the sequence and then reappear (in this case 1000 lines later). This degradation is more clear in perplexity as shown in Figure \ref{fig:sensitivity_perplexity} thus we choose it as our main anomaly detection metric.

\begin{figure}[htbp]
\centering

\begin{subfigure}[t]{0.49\linewidth}
    \centering
    \includegraphics[width=\linewidth]{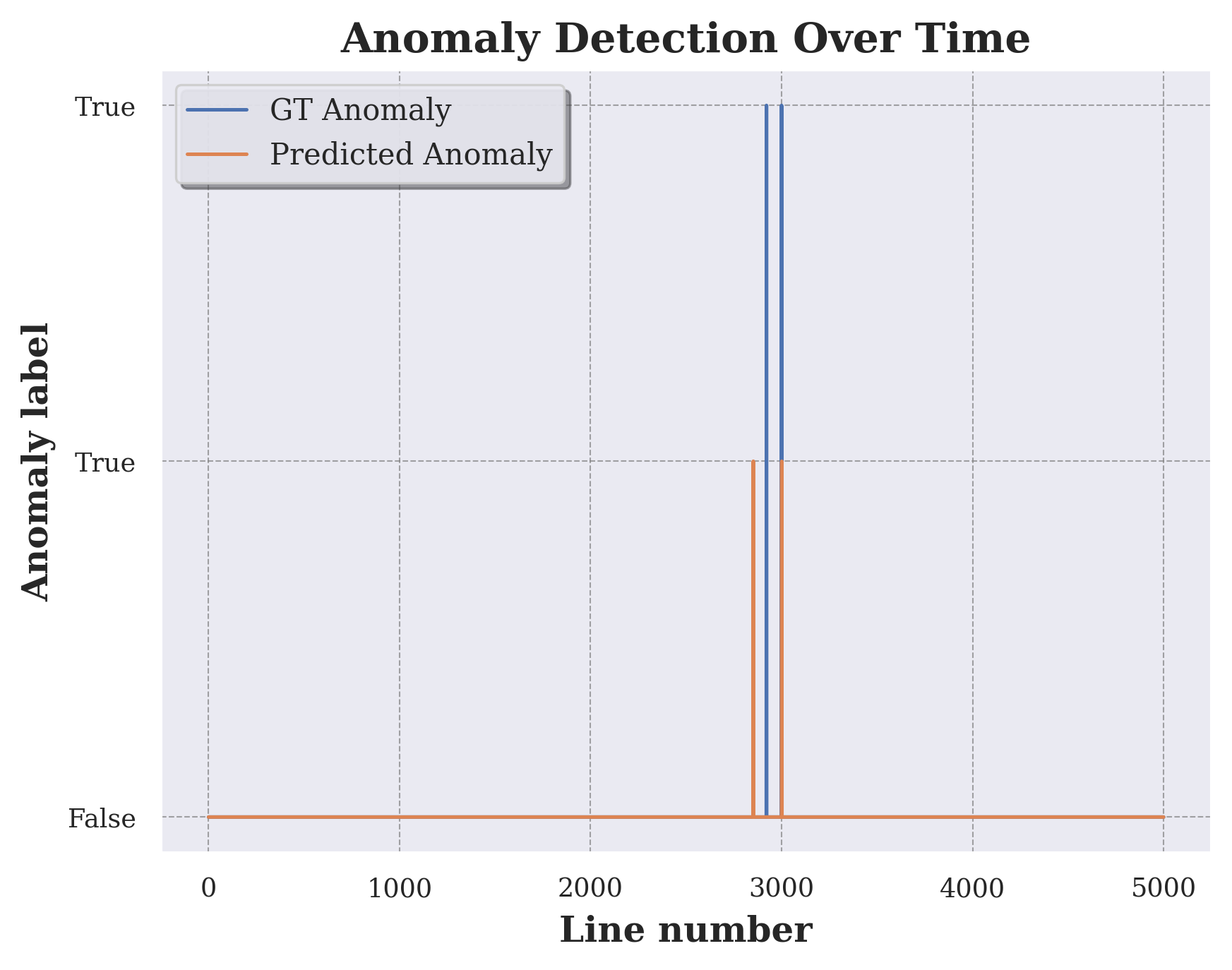}
    \caption{Normal operation - Sample 1}
    \label{fig:normal_op_1}
\end{subfigure}
\hfill
\begin{subfigure}[t]{0.49\linewidth}
    \centering
    \includegraphics[width=\linewidth]{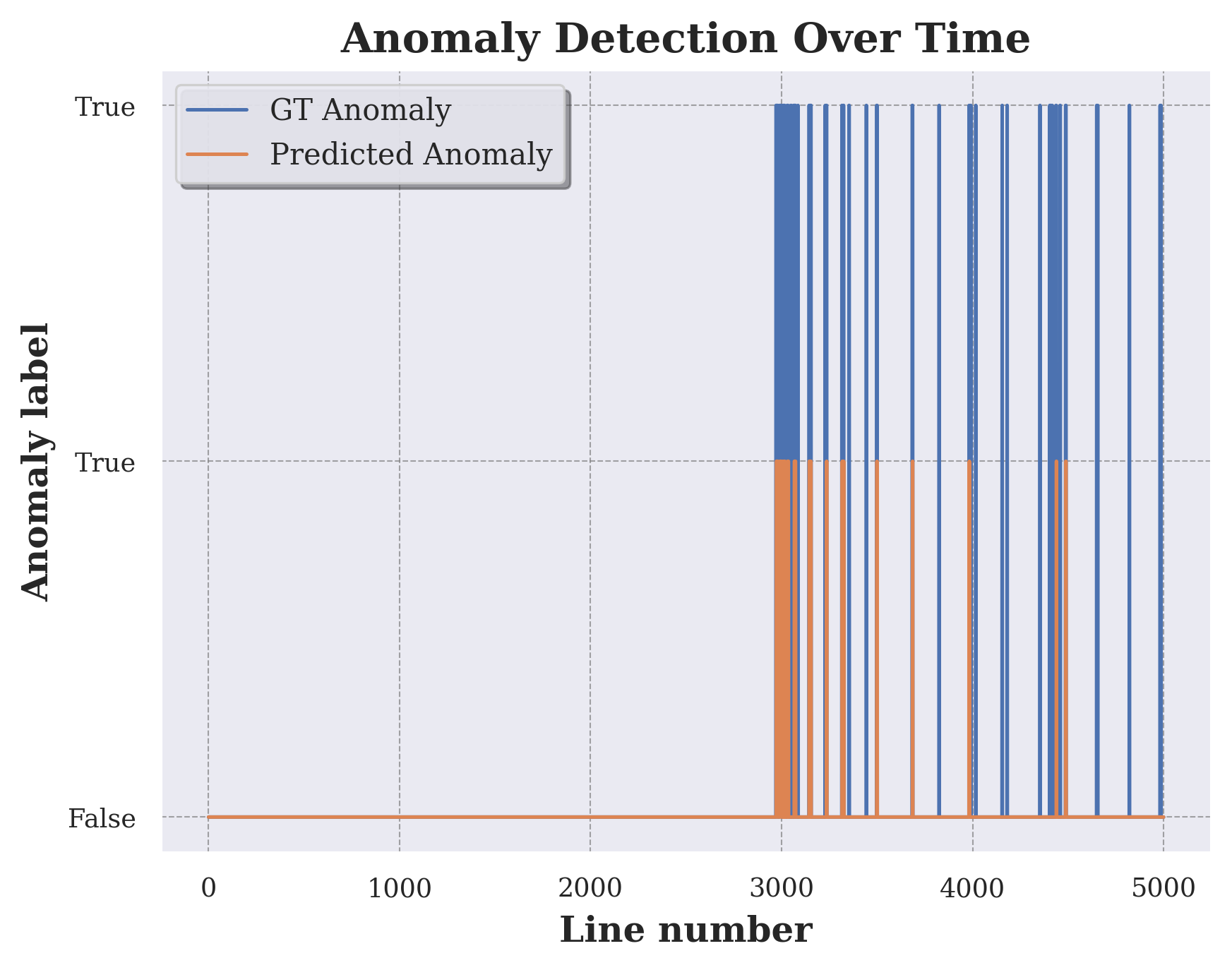}
    \caption{System restart - Sample 1}
    \label{fig:system_restart_1}
\end{subfigure}
    
\vspace{1em}

\begin{subfigure}[t]{0.49\linewidth}
    \centering
    \includegraphics[width=\linewidth]{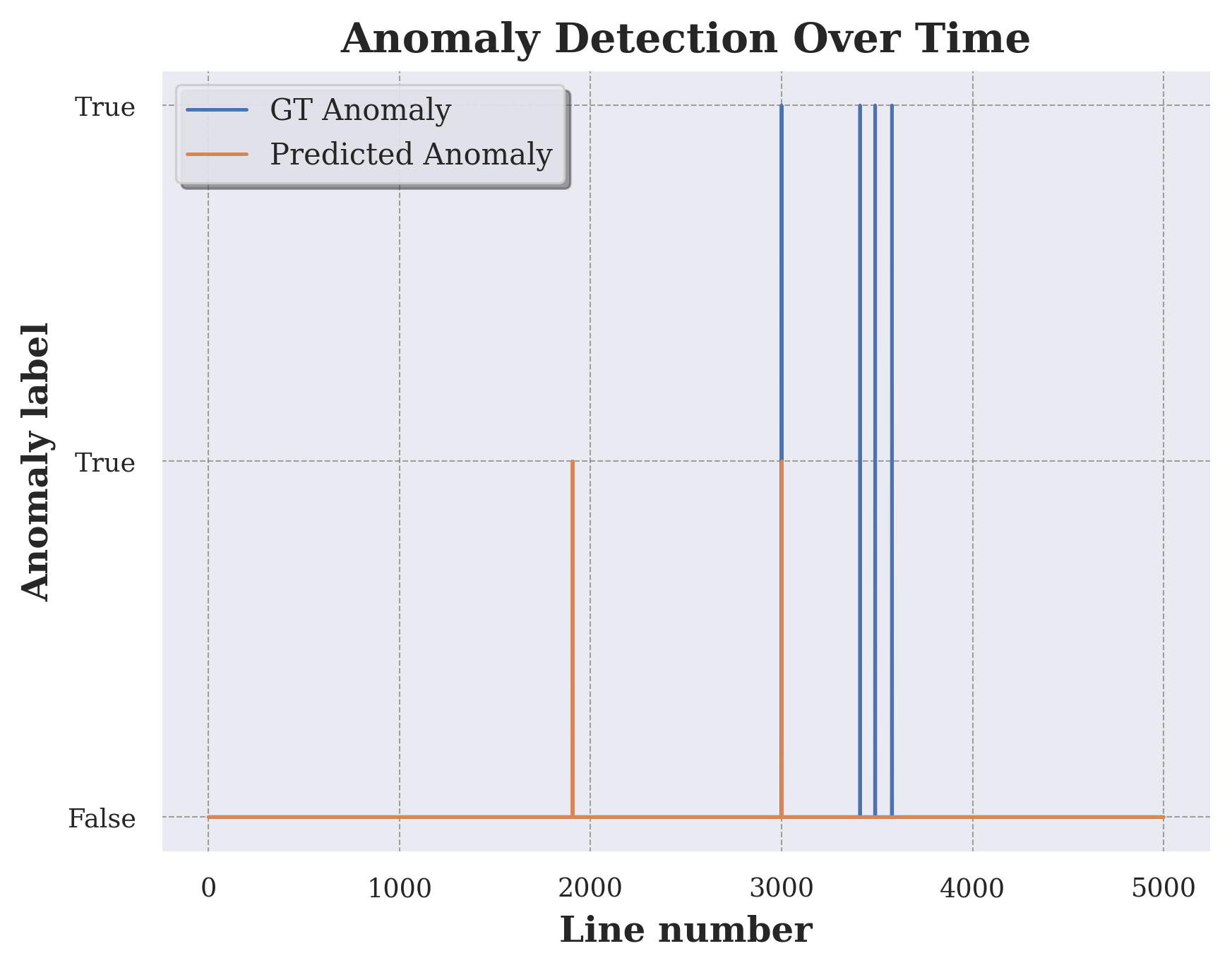}
    \caption{Normal operation - Sample 2}
    \label{fig:normal_op_2}
\end{subfigure}
\hfill
\begin{subfigure}[t]{0.49\linewidth}
    \centering
    \includegraphics[width=\linewidth]{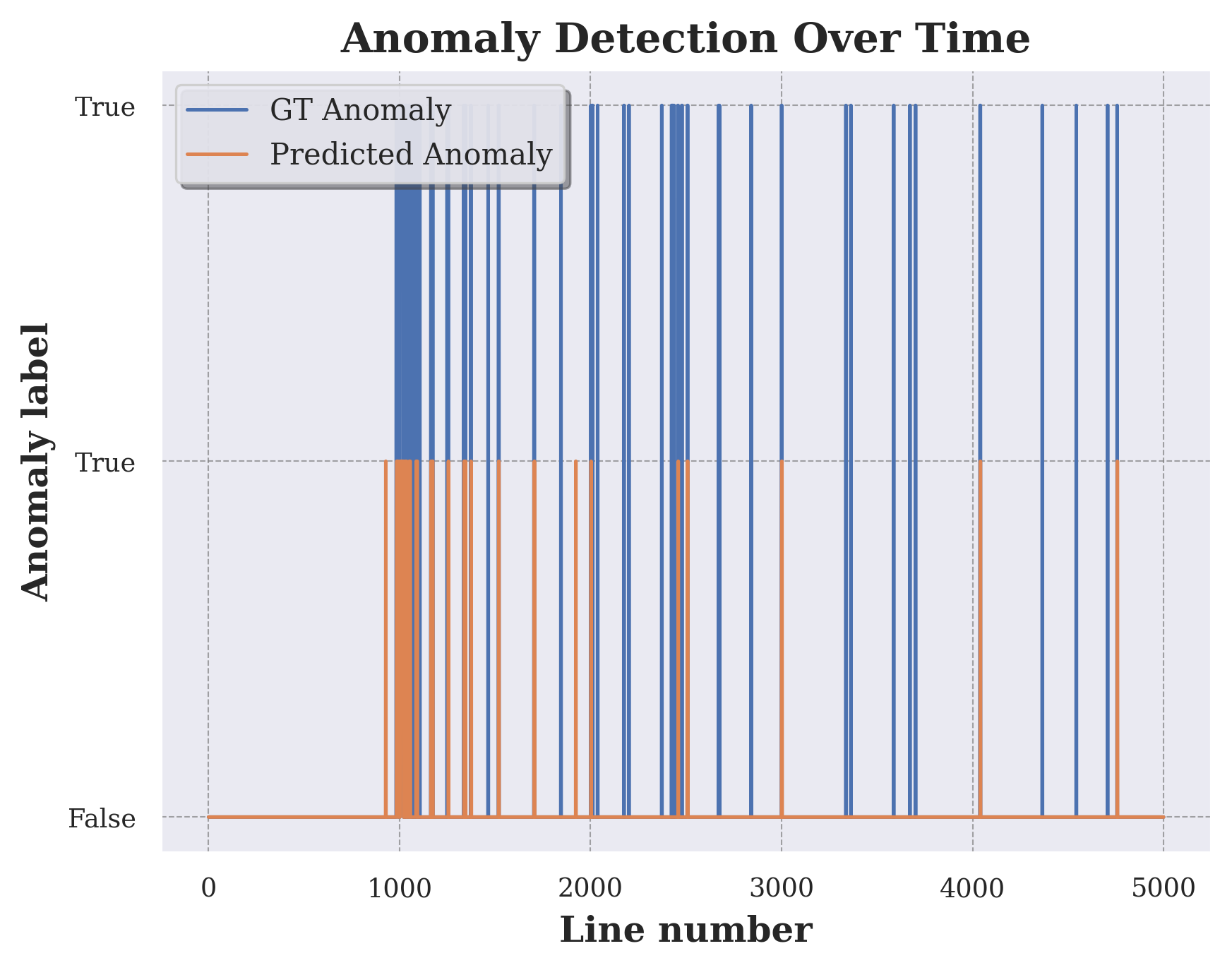}
    \caption{System restart - Sample 2}
    \label{fig:system_restart_2}
\end{subfigure}

\caption{Visualization of detected anomalies. Orange lines represent model predictions, while blue lines depicted with an offset represent actual anomalies. Left column (a, c) corresponds to normal operation scenarios, and the right column (b, d) corresponds to test scenarios involving a system restart.}
\label{fig:comparison_visual_ad}
\end{figure}

Next we visualize how our model performs in comparison to the traditional communication analysis system. Specifically we now look at the artifact of phase 2 - fine-tuning. Phase 2 further boosts our detection performance from phase 1 and also accounts for the inconsistent labeling process. The labeling process produces two types of test scenarios: 1) scenarios in which the system is restarted illustrated in Figure \ref{fig:system_restart_1} and Figure \ref{fig:system_restart_2} to force the occurrence of frequent anomalies and scenarios in which the system is already stable and anomalies are less frequent as shown in Figure \ref{fig:normal_op_1} and Figure \ref{fig:normal_op_2}. In both scenarios, our solution is able to successfully identify the most important anomalies.

Fine-tuning was performed using LoRA \cite{hu2021loralowrankadaptationlarge} with a rank of $64$, learning rate of $5\mathrm{e}{-5}$ and $\alpha$ of $0.5$. The detections were computed based on perplexity. We found out experimentally that top-$k$ based detections result in lower overall recall on the test dataset. At the same time we used a filter width of $3$ consecutive lines and a filter threshold of $1.5$.

%% file: chapters/conclusion.tex
\section{Conclusion}
In this paper, we presented a Large Language Model for detecting abnormal ECU logs, built upon data labeled by an unreliable algorithm. The results suggest that having a model first trained to learn the ECU language and in a second step, making the model behave similarly for the same entries, yields the best results in our circumstances. 

This proves that with the least amount of correct labels, the model is able to generalize, without blindly following the inconsistent behavior of the annotator, but only the part of it that makes most sense or is more consistent.

There are other ideas as well worth experimenting within the current context, such as: clustering, supervised methods (e.g. label smoothing) or semi-supervised methods. In terms of model architecture, an alternative to decoder-only methods would be the encoder-only ones, such as BERT \cite{devlin2019bertpretrainingdeepbidirectional}.

Our vision would be to create a framework that can be adopted in other domain-specific environments as well, without the need of specific labels for a certain use case, but rather some clues.

%% file: root.bbl
\begin{thebibliography}{10}
\providecommand{\url}[1]{#1}
\csname url@rmstyle\endcsname
\providecommand{\newblock}{\relax}
\providecommand{\bibinfo}[2]{#2}
\providecommand\BIBentrySTDinterwordspacing{\spaceskip=0pt\relax}
\providecommand\BIBentryALTinterwordstretchfactor{4}
\providecommand\BIBentryALTinterwordspacing{\spaceskip=\fontdimen2\font plus
\BIBentryALTinterwordstretchfactor\fontdimen3\font minus \fontdimen4\font\relax}
\providecommand\BIBforeignlanguage[2]{{%
\expandafter\ifx\csname l@#1\endcsname\relax
\typeout{** WARNING: IEEEtran.bst: No hyphenation pattern has been}%
\typeout{** loaded for the language `#1'. Using the pattern for}%
\typeout{** the default language instead.}%
\else
\language=\csname l@#1\endcsname
\fi
#2}}

\bibitem{su2024largelanguagemodelsforecasting}
\BIBentryALTinterwordspacing
J.~Su, C.~Jiang, X.~Jin, Y.~Qiao, T.~Xiao, H.~Ma, R.~Wei, Z.~Jing, J.~Xu, and J.~Lin, ``Large language models for forecasting and anomaly detection: A systematic literature review,'' 2024. [Online]. Available: \url{https://arxiv.org/abs/2402.10350}
\BIBentrySTDinterwordspacing

\bibitem{devlin2019bertpretrainingdeepbidirectional}
\BIBentryALTinterwordspacing
J.~Devlin, M.-W. Chang, K.~Lee, and K.~Toutanova, ``Bert: Pre-training of deep bidirectional transformers for language understanding,'' 2019. [Online]. Available: \url{https://arxiv.org/abs/1810.04805}
\BIBentrySTDinterwordspacing

\bibitem{guo2021logbertloganomalydetection}
\BIBentryALTinterwordspacing
H.~Guo, S.~Yuan, and X.~Wu, ``Logbert: Log anomaly detection via bert,'' 2021. [Online]. Available: \url{https://arxiv.org/abs/2103.04475}
\BIBentrySTDinterwordspacing

\bibitem{lee2023lanobertloganomalydetection}
\BIBentryALTinterwordspacing
Y.~Lee, J.~Kim, and P.~Kang, ``Lanobert: System log anomaly detection based on bert masked language model,'' 2023. [Online]. Available: \url{https://arxiv.org/abs/2111.09564}
\BIBentrySTDinterwordspacing

\bibitem{alkhatib2022canbertitcontrollerarea}
\BIBentryALTinterwordspacing
N.~Alkhatib, M.~Mushtaq, H.~Ghauch, and J.-L. Danger, ``Can-bert do it? controller area network intrusion detection system based on bert language model,'' 2022. [Online]. Available: \url{https://arxiv.org/abs/2210.09439}
\BIBentrySTDinterwordspacing

\bibitem{zhao2024weaklysupervisedanomalydetection}
\BIBentryALTinterwordspacing
H.~Zhao, C.~Zi, Y.~Liu, C.~Zhang, Y.~Zhou, and J.~Li, ``Weakly supervised anomaly detection via knowledge-data alignment,'' 2024. [Online]. Available: \url{https://arxiv.org/abs/2402.03785}
\BIBentrySTDinterwordspacing

\bibitem{das2023fewshotanomalydetectiontext}
\BIBentryALTinterwordspacing
A.~S. Das, A.~Ajay, S.~Saha, and M.~Bhuyan, ``Few-shot anomaly detection in text with deviation learning,'' 2023. [Online]. Available: \url{https://arxiv.org/abs/2308.11780}
\BIBentrySTDinterwordspacing

\bibitem{sennrich2016neuralmachinetranslationrare}
\BIBentryALTinterwordspacing
R.~Sennrich, B.~Haddow, and A.~Birch, ``Neural machine translation of rare words with subword units,'' 2016. [Online]. Available: \url{https://arxiv.org/abs/1508.07909}
\BIBentrySTDinterwordspacing

\bibitem{Radford2019LanguageMA}
\BIBentryALTinterwordspacing
A.~Radford, J.~Wu, R.~Child, D.~Luan, D.~Amodei, and I.~Sutskever, ``Language models are unsupervised multitask learners,'' 2019. [Online]. Available: \url{https://api.semanticscholar.org/CorpusID:160025533}
\BIBentrySTDinterwordspacing

\bibitem{yang2024qwen2technicalreport}
\BIBentryALTinterwordspacing
A.~Yang, B.~Yang, B.~Hui, B.~Zheng, B.~Yu, C.~Zhou, C.~Li, C.~Li, D.~Liu, F.~Huang, G.~Dong, H.~Wei, H.~Lin, J.~Tang, J.~Wang, J.~Yang, J.~Tu, J.~Zhang, J.~Ma, J.~Yang, J.~Xu, J.~Zhou, J.~Bai, J.~He, J.~Lin, K.~Dang, K.~Lu, K.~Chen, K.~Yang, M.~Li, M.~Xue, N.~Ni, P.~Zhang, P.~Wang, R.~Peng, R.~Men, R.~Gao, R.~Lin, S.~Wang, S.~Bai, S.~Tan, T.~Zhu, T.~Li, T.~Liu, W.~Ge, X.~Deng, X.~Zhou, X.~Ren, X.~Zhang, X.~Wei, X.~Ren, X.~Liu, Y.~Fan, Y.~Yao, Y.~Zhang, Y.~Wan, Y.~Chu, Y.~Liu, Z.~Cui, Z.~Zhang, Z.~Guo, and Z.~Fan, ``Qwen2 technical report,'' 2024. [Online]. Available: \url{https://arxiv.org/abs/2407.10671}
\BIBentrySTDinterwordspacing

\bibitem{DBLP:journals/corr/abs-2004-10964}
\BIBentryALTinterwordspacing
S.~Gururangan, A.~Marasovic, S.~Swayamdipta, K.~Lo, I.~Beltagy, D.~Downey, and N.~A. Smith, ``Don't stop pretraining: Adapt language models to domains and tasks,'' \emph{CoRR}, vol. abs/2004.10964, 2020. [Online]. Available: \url{https://arxiv.org/abs/2004.10964}
\BIBentrySTDinterwordspacing

\bibitem{hu2021loralowrankadaptationlarge}
\BIBentryALTinterwordspacing
E.~J. Hu, Y.~Shen, P.~Wallis, Z.~Allen-Zhu, Y.~Li, S.~Wang, L.~Wang, and W.~Chen, ``Lora: Low-rank adaptation of large language models,'' 2021. [Online]. Available: \url{https://arxiv.org/abs/2106.09685}
\BIBentrySTDinterwordspacing

\bibitem{gu2024mambalineartimesequencemodeling}
\BIBentryALTinterwordspacing
A.~Gu and T.~Dao, ``Mamba: Linear-time sequence modeling with selective state spaces,'' 2024. [Online]. Available: \url{https://arxiv.org/abs/2312.00752}
\BIBentrySTDinterwordspacing

\bibitem{hoffmann2022trainingcomputeoptimallargelanguage}
\BIBentryALTinterwordspacing
J.~Hoffmann, S.~Borgeaud, A.~Mensch, E.~Buchatskaya, T.~Cai, E.~Rutherford, D.~de~Las~Casas, L.~A. Hendricks, J.~Welbl, A.~Clark, T.~Hennigan, E.~Noland, K.~Millican, G.~van~den Driessche, B.~Damoc, A.~Guy, S.~Osindero, K.~Simonyan, E.~Elsen, J.~W. Rae, O.~Vinyals, and L.~Sifre, ``Training compute-optimal large language models,'' 2022. [Online]. Available: \url{https://arxiv.org/abs/2203.15556}
\BIBentrySTDinterwordspacing

\end{thebibliography}
